# Localization of Internet-based Mobile Robot

**Manh Duong Phung, Thi Thanh Van Nguyen, Thuan Hoang Tran, Quang Vinh Tran**

*University of Engineering and Technology, Vietnam National University, Hanoi*
*144 Xuan Thuy, Cau Giay, Hanoi, Vietnam*

Corresponding author*: duongpm@vnu.edu.vn*

**ABSTRACT**

This paper presents a new optimal filter namely past observation-based extended Kalman filter for the problem of localization of Internet-based mobile robot in which the control input and the feedback measurement suffer from communication delay. The filter operates through two phases: the time update and the data correction. The time update predicts the robot position by reformulating the kinematics model to be non-memoryless. The correction step corrects the prediction by extrapolating the delayed measurement to the present and then incorporating it to the being estimate as there is no delay. The optimality of the incorporation is ensured by the derivation of a multiplier that reflects the relevance of past observations to the present. Simulations in MATLAB and experiments in a real networked robot system confirm the validity of the proposed approach.

## I. INTRODUCTION

Internet-based robotic systems are attracting more research attention due to their ability to open new applications from home service such as remote cleaning to industrial manufacture like tele-manipulator. Whereas early works tried to answer the question of how to control a robot through the Internet [1, 2], recent researches focused on controlling it in real time and dealing with advanced problems such as map matching, path following, and point-to-point stabilization when controlling via the Internet becomes more easily with the support of the embedded Ethernet, on-chip web server, scripting language, socket programming, etc. [3-6].

In this paper, the localization problem is addressed. Localization, that is the estimation of robot's location and orientation from sensor data, is a major problem in mobile robotics. In order to autonomously accomplish given tasks, the robot need to know its position and orientation relative to a pre-defined frame. Various approaches have been proposed and significant progresses have been made on this front such as the dead reckoning, the map matching, the landmark, and the Bayesian methods [7-10].

When operating over the Internet, the localization however poses several new challenges. They are the inevitable communication delays, the out-of-sequence data arrival and the partial intermittent. To the author's knowledge, only few works addressed these problems and often in an indirect way. In [11], the robot pose is estimated at the local site using the sensor system. The information is then transmitted to the remote site as the robot pose at the receiving time without considering the change of the robot during the communication. In [12], a global map which is proportional to the real dimension of a laboratory is constructed in the client side. The absolute position of the robot is then determined by comparing this global map with a reference map of the local site. A posture estimator is proposed in [13]. It employs the robot state such as wheels' velocities and the network parameters such as the time delay to predict the robot position and orientation. It is recognizable from the proposed approaches that the key issue of control over the Internet, the communication channel, is avoided to

Corresponding author: Phung Manh Duong, VNU University of Engineering and Techonology
Address: 315 G2 Building, 144 Xuan Thuy, Cau Giay
Email: duongpm@vnu.edu.vn – Mobile phone: 0983 361 683 – Tel: 04 37549272

cope with. The data transmission between the remote controller and the actuator was treated as a given condition and rarely touched. From the viewpoint of control theory, significant delay is equivalent to inaccuracy in state estimation and control that can easily downgrade the system performance.

In this paper, the localization problem of Internet-based mobile robot is investigated. The Internet introduces latency to the data exchanged between the remote controller and the local actuator and sensor. A novel filter namely past observation-based extended Kalman filter (PO-EKF) inspired from a well-known optimal filter, the Kalman filter, is proposed. This filter enables the incorporation of delayed measurements to the *posteriori* estimation by introducing a "relevance factor" which describes the relevance of observations from the past to the present. Simulations have been carried out in MATLAB and experiments have been implemented in a real Internet-based mobile robot system. The results confirm the effectiveness of the proposed approach.

The paper is arranged as follows. Details of the localization problem are described in Section II. The algorithm for state estimation using the PO-EKF is explained in Section III. Section IV introduces the simulations and experiments. The paper concludes with an evaluation of the system.

## II. SYSTEM MODELLING AND PROBLEM FORMULATION

In this paper, the two wheeled, differential-drive mobile robot with non-slipping and pure rolling is considered. The state of the robot is the position of the wheels axis center (x, y) and the chassis orientation $\theta$ with respect to the x axis. Figure 1 shows the coordinate systems and notations for the robot where $R$ denotes the radius of driven wheels and $L$ denotes the distance between the wheels.

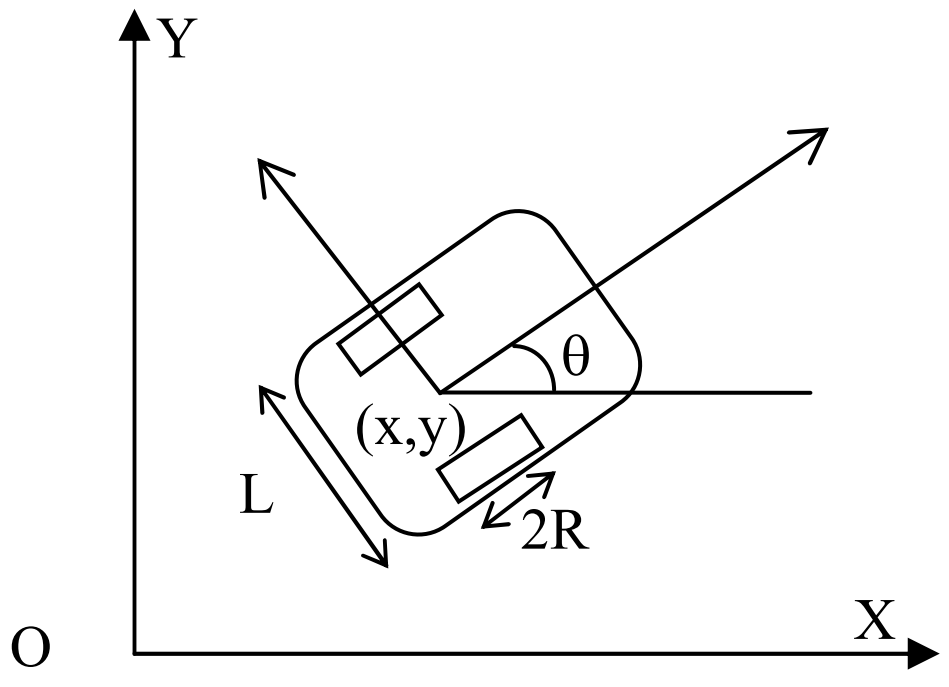

Figure 1: The robot's pose and parameters

Let $T_s$ be the sampling period, $\omega_L(k)$ and $\omega_R(k)$ be the measurements of rotational speed of the left and right wheels with the encoders at the time $k$, respectively. The discrete kinematics model of the robot is given by:

$$\begin{aligned} x_{k+1} &= x_k + \frac{R}{2}T_s(\omega_L(k)+\omega_R(k))\cos\theta_k \\ y_{k+1} &= y_k + \frac{R}{2}T_s(\omega_L(k)+\omega_R(k))\sin\theta_k \\ \theta_{k+1} &= \theta_k + \frac{R}{L}T_s(\omega_L(k)-\omega_R(k)) \end{aligned} \tag{1}$$

In practice, (1) suffers from unavoidable errors appeared in the system. Errors can be both systematic such as the imperfectness of robot model and nonsystematic such as the slip of wheels. These errors have accumulative characteristic so that they can break the stability of the system if appropriate compensation is not considered. In order to capture these scenarios, the system model is rewritten in the state space representation as follows.

Let $\mathbf{x} = [x\, y\, \theta]^T$ be the state vector. This state can be observed by an absolute measurement, $\mathbf{z}$. This measurement is described by a nonlinear function, $h$, of the robot coordinates and a noise process, $\mathbf{v}$.

Denoting the function (1) as $f$, with an input vector $\mathbf{u}$ and a disturbance $\mathbf{w}$, the robot is then described by:

$$\begin{aligned} \mathbf{x}_{k+1} &= f_k(\mathbf{x}_k, \mathbf{u}_k, \mathbf{w}_k) \\ \mathbf{z}_k &= h_k(\mathbf{x}_k, \mathbf{v}_k) \end{aligned} \tag{2}$$

where the random variables $\mathbf{w}_k$ and $\mathbf{v}_k$ represent the process and measurement noises respectively. They are assumed to be independent to each other, white, and with normal probability distributions: $\mathbf{w}_k \sim \mathbf{N}(0, Q_k)$ $\mathbf{v}_k \sim \mathbf{N}(0, R_k)$ $E(\mathbf{w}_i \mathbf{v}_j^T) = 0$

Now, consider the robot system when distributing over the Internet. The system becomes decentralized and its functioning operation depends on a number of network parameters such as the time delay, the data loss and the out-of-order data arrival. Among parameters, the time delay introduces the dominant influence and is focused in this paper. Other issues will be addressed in future work. The operation of the system can be described in Figure 2.

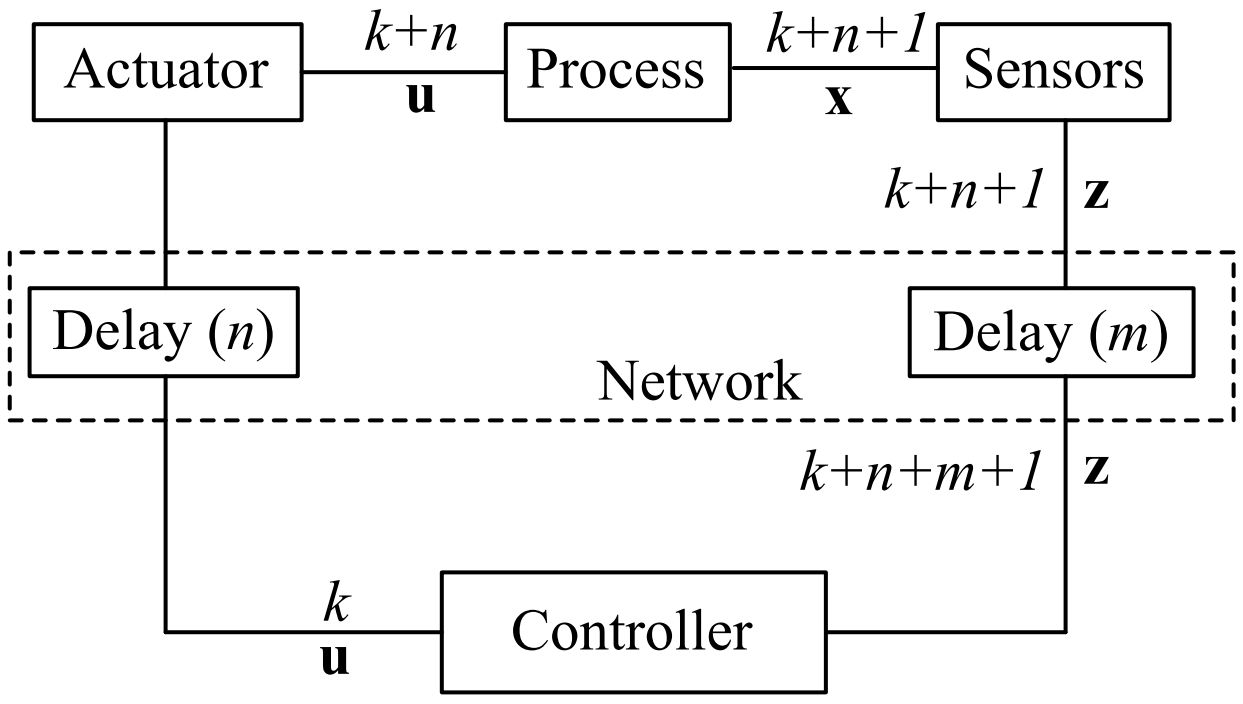

Figure 2: Model of Internet-based robot system

At time $k$, the controller sends a control input $\mathbf{u}$ to the actuator. Due to the network delay $n$, the control signal arrives to the actuator at time $k+n$. After one sampling period to the time $k+n+1$, the system state changes and the sensor updates this by taking a measurement $\mathbf{z}$. The measurement is transmitted over the Internet to the state estimator at time $k+n+m+1$ in which $m$ is the time delay of the transmission. The measurement $\mathbf{z}$, the control input $\mathbf{u}$, and the knowledge of the system are then incorporated in the estimator to extract the state of the system. This estimate is employed as the feedback for the controller to start a new control cycle.

From the analysis, the robot state at time $k$ is driven by the control input at time $k-n-1$ while the state estimation at time $k$ is actually based on the measurement at time $k-m$. The system is therefore non-memoryless and can be rewritten as follows:

$$\begin{aligned} \mathbf{x}_k &= f_{k-1}(\mathbf{x}_{k-1}, \mathbf{u}_{k-n-1}, \mathbf{w}_{k-1}) \\ \mathbf{z}_k^i &= h_{k-m}(\mathbf{x}_{k-m}, \mathbf{v}_{k-m}) \end{aligned} \tag{3}$$

where $i=k-m$ is the time that the delayed measurement $\mathbf{z}_k^i$ is taken. Our approach for the problem of localization is the development of an algorithm for the no-Internet case and then extends it to cope with network problems.

## III. LOCALIZATION ALGORITHM

As defined in previous section, the state vector $\mathbf{x}_k$ includes the position and orientation of the mobile robot. Consequently, the robot localization becomes the problem of state estimation with the state space model described in (3). The optimal solution for this is the Kalman filter [14, 15]. In this section, the standard Kalman filter is first briefly described. The past observation-based Kalman filter (PO-KF) is then derived to cope with the time delay induced by the Internet. Finally, the PO-KF is extended to apply to the nonlinear robot system.

### *1. The standard Kalman filter*

The Kalman filter by definition is a set of mathematical equations that provides an efficient computational (recursive) means to estimate the state of a process, in a way that minimizes the mean of the squared error [15]. Consider a linear discrete system observed by measurements in which both are subjected to noises as follows:

$$\begin{aligned} \mathbf{x}_k &= A_{k-1}\mathbf{x}_{k-1} + B_{k-1}\mathbf{u}_{k-1} + \mathbf{w}_{k-1} \\ \mathbf{z}_k &= H_k\mathbf{x}_k + \mathbf{v}_k \end{aligned} \tag{4}$$

The noises $\mathbf{w}_k$ and $\mathbf{v}_k$ are assumed to be independent to each other, white, and with normal probability distributions: $\mathbf{w}_k \sim \mathbf{N}(0,Q_k) \quad \mathbf{v}_k \sim \mathbf{N}(0,R_k) \quad E(\mathbf{w}_i\mathbf{v}_j^{\ T}) = 0$

The steps to calculate the Kalman filter for the system (4) can be summarized as follows:

- The time update equations (prediction phase):

$$\begin{aligned} \hat{\mathbf{x}}_k^- &= A_{k-1}\hat{\mathbf{x}}_{k-1}^+ + B_{k-1}\mathbf{u}_{k-1} \\ P_k^- &= A_{k-1}P_{k-1}^+A_{k-1}^T + Q_{k-1} \end{aligned} \tag{5}$$

  where $\hat{\mathbf{x}}_k^- \in \Re^n$ is the *priori* state estimate at step *k* given knowledge of the process prior to step *k*, $P_k^-$ denotes the covariance matrix of the state-prediction error and $Q_{k-1}$ is the input-noise covariance matrix.

- The data update equations (correction phase):

$$\begin{aligned} K_k &= P_k^- H_k^T [H_k P_k^- H_k^T + R_k]^{-1} \\ \hat{\mathbf{x}}_k^+ &= \hat{\mathbf{x}}_k^- + K_k[\mathbf{z}_k - H_k\hat{\mathbf{x}}_k^-] \\ P_k^+ &= [I - K_kH_k]P_k \end{aligned} \tag{6}$$

  where $\hat{\mathbf{x}}_k \in \Re^n$ is the *posteriori* state estimate at step *k* given measurement $\mathbf{z}_k$, $K_k$ is the Kalman gain and $R_k$ is the covariance matrix of measurement noise.

When operating over the Internet, both control inputs and observation measurements suffer from the communication delay. This delayed information cannot be fused using the standard Kalman filter but requires some modifications in the structure of the filter.

### *2. The past observation-based Kalman filter with time delay*

From (3), the state $\mathbf{x}_k$ at time *k* actually reflects the effect of the input $\mathbf{u}$ at time *k-n-1*. The time update equation (5) of the Kalman filter therefore can be rewritten as follows:

$$\hat{\mathbf{x}}_k^- = A_{k-1}\hat{\mathbf{x}}_{k-1}^+ + B_{k-n-1}\mathbf{u}_{k-n-1} \tag{7}$$

In order to derive the new data update equation, we consider the measurement $\mathbf{z}_k^i$ in (3). This measurement was taken at the previous time *i*. Due to the delay, it could not reach the estimator until time *k*. Therefore, we construct the data update equation of the form:

$$\hat{\mathbf{x}}_k^+ = \hat{\mathbf{x}}_k^- + K_k(\mathbf{z}_k^i - H_i\hat{\mathbf{x}}_i^-) \tag{8}$$

and recompute the Kalman gain and error covariance to ensure the optimality of the new equation.

**Kalman gain and Posteriori Error Covariance**: Assume that the measurement is fused using (8) with an arbitrary gain $K_k$. The covariance of the *posteriori* estimate error, $P_k^+$, is determined as:

$$
\begin{aligned}
P_k^+ &= E(\mathbf{e}_k^+\mathbf{e}_k^{+T}) \\
&= E[\mathbf{e}_k^-\mathbf{e}_k^{-T} - \mathbf{e}_k^-\mathbf{e}_i^{-T}(K_kH_i)^T - \mathbf{e}_k^-\mathbf{v}_i^TK_k^T - K_kH_i\mathbf{e}_i^-\mathbf{e}_k^{-T} \\
&\quad + K_kH_i\mathbf{e}_i^-\mathbf{e}_i^{-T}(K_kH_i)^T + K_kH_i\mathbf{e}_i^-\mathbf{v}_i^TK_k^T - K_k\mathbf{v}_i\mathbf{e}_k^{-T} \\
&\quad + K_k\mathbf{v}_i\mathbf{e}_i^{-T}(K_kH_i)^T + K_k\mathbf{v}_i\mathbf{v}_i^TK_k^T]
\end{aligned} \tag{9}
$$

Due to the independence between $\mathbf{e}^-$ and $\mathbf{v}$, (9) can be simplified to:

$$
\begin{aligned}
P_k^+ &= E(\mathbf{e}_k^-\mathbf{e}_k^{-T}) - E(\mathbf{e}_k^-\mathbf{e}_i^{-T})(K_kH_i)^T - K_kH_iE(\mathbf{e}_i^-\mathbf{e}_k^{-T}) \\
&\quad + K_kH_iE(\mathbf{e}_i^-\mathbf{e}_i^{-T})(K_kH_i)^T + K_kE(\mathbf{v}_i\mathbf{v}_i^T)K_k^T] \\
&= P_k^- - L^TH_i^TK_k^T - K_kH_iL + K_kH_iP_i^-H_i^TK_k^T + K_kR_iK_k^T
\end{aligned} \tag{10}
$$

where $L = E(\mathbf{e}_i^-\mathbf{e}_k^{-T})$.

As the matrix $K_k$ is chosen to be the gain or blending factor that minimizes the *posteriori* error covariance, this minimization is accomplished by taking the derivative of the trace of the *posteriori* error covariance with respect to $K_k$, setting that result equal to zero, and then solving for $K_k$. Applying this process to (10) obtains:

$$
\frac{\partial tr(P_k^+)}{\partial K_k} = 2(-L^TH_i^T + K_kH_iP_iH_i^T + K_kR_i) = 0 \tag{11}
$$

$$
\Leftrightarrow K_k = L^TH_i^T[H_iP_i^-H_i^T + \tilde{R}_i]^{-1} \tag{12}
$$

Inserting (12) in (10) leads to a simpler form of $P_k^+$:

$$
P_k^+ = P_k^- - K_kH_iL \tag{13}
$$

In order to compute $L$, the *priori* state estimate at time $k$ needs determining from the estimate at time $i$. Through the time update (7) and the data update (8), $\mathbf{e}^-$ becomes:

$$
\begin{aligned}
\mathbf{e}_k^- &= \mathbf{x}_k - \hat{\mathbf{x}}_k^- \\
&= A_{k-1}\mathbf{e}_{k-1}^+ - \mathbf{w}_{k-1} \\
&= A_{k-1}[(I - K_{k-1}H_{k-1})\mathbf{e}_{k-1}^- + K_{k-1}\mathbf{v}_{k-1}] - \mathbf{w}_{k-1}
\end{aligned} \tag{14}
$$

After $m$ updating steps, the estimation error becomes:

$$
\mathbf{e}_k^- = F\mathbf{e}_i^- + \xi_1(\mathbf{w}_i,...,\mathbf{w}_{k-1}) + \xi_2(\mathbf{v}_i,...,\mathbf{v}_{k-1}) \tag{15}
$$

where

$$
F = \prod_{j=1}^{m} A_{k-j}(I - K_{k-j}H_{k-j}) \tag{16}
$$

and $\xi_1$ and $\xi_2$ are the functions of noise sequences $\mathbf{w}$ and $\mathbf{v}$. From (15) and the independence between $\mathbf{e}^-$ and noise sequences, the covariance $L$ becomes:

$$
L = E(\mathbf{e}_i^-\mathbf{e}_k^{-T}) = P_i^-F^T \tag{17}
$$

Substituting (17) in (13) and (12) yields:

$$
P_k^+ = P_k^- - K_kH_iP_i^-F^T \tag{18}
$$

and

$$
K_k = FP_i^-H_i^T[H_iP_i^-H_i^T + R_i]^{-1} \tag{19}
$$

The PO-KF for the networked linear system can be summarized as follows:

- The time update equations:

$$\begin{aligned} \hat{\mathbf{x}}_k^- &= A_{k-1}\hat{\mathbf{x}}_{k-1}^+ + B_{k-1}\mathbf{u}_{k-n-1} \\ P_k^- &= A_{k-1}P_{k-1}^+A_{k-1}^T + Q_{k-1} \end{aligned} \tag{20}$$

- The data update equations:

$$\begin{aligned} M_* &= \prod_{i=1}^{m} A_{k-i}(I - K_{k-i}H_{k-i}) \\ K_k &= M_* P_s^- H_s^T [H_s P_s^- H_s^T + R_s]^{-1} \\ \hat{\mathbf{x}}_k^+ &= \hat{\mathbf{x}}_k^- + K_k[\mathbf{z}_k^* - H_s\hat{\mathbf{x}}_s^-] \\ P_k^+ &= P_k^- - K_k H_k P_s^- M_*^T \end{aligned} \tag{21}$$

### *3. The past observation-based extended Kalman filter for Internet-based robot systems*

Though the PO-KF derived in previous section is capable for networked control systems, the system has to be linear. As our robot system is nonlinear, further modification needs to be accomplished. In this section, the derivation of the EKF is inherited to extend the PO-KF for the nonlinear system. The main idea is the linearization of a nonlinear system around its previous estimated states.
Performing a Taylor series expansion of the state equation around $(\hat{\mathbf{x}}_{k-1}^+, \mathbf{u}_{k-1}, 0)$ gives:

$$\begin{aligned} \mathbf{x}_k &= f_{k-1}(\hat{\mathbf{x}}_{k-1}^+, \mathbf{u}_{k-1}, 0) + \left.\frac{\partial f_{k-1}}{\partial \mathbf{x}}\right|_{(\hat{\mathbf{x}}_{k-1}^+, \mathbf{u}_{k-1}, 0)} (\mathbf{x}_{k-1} - \hat{\mathbf{x}}_{k-1}^+) + \left.\frac{\partial f_{k-1}}{\partial \mathbf{w}}\right|_{(\hat{\mathbf{x}}_{k-1}^+, \mathbf{u}_{k-1}, 0)} \mathbf{w}_{k-1} \\ &= f_{k-1}(\hat{\mathbf{x}}_{k-1}^+, \mathbf{u}_{k-1}, 0) + A_{k-1}(\mathbf{x}_{k-1} - \hat{\mathbf{x}}_{k-1}^+) + W_{k-1}\mathbf{w}_{k-1} \\ &= A_{k-1}\mathbf{x}_{k-1} + [f_{k-1}(\hat{\mathbf{x}}_{k-1}^+, \mathbf{u}_{k-1}, 0) - A_{k-1}\hat{\mathbf{x}}_{k-1}^+] + W_{k-1}\mathbf{w}_{k-1} \\ &= A_{k-1}\mathbf{x}_{k-1} + \tilde{\mathbf{u}}_{k-1} + \tilde{\mathbf{w}}_{k-1} \end{aligned} \tag{22}$$

where$A_{k-1}$,$W_{k-1}$, $\tilde{\mathbf{u}}_{k-1}$, $\tilde{\mathbf{w}}_{k-1}$ are defined by the above equation. Similarly, the measurement equation is linearized around $\mathbf{x}_k = \hat{\mathbf{x}}_k^-$ and $\mathbf{v}_k = 0$ to obtain

$$\begin{aligned} \mathbf{z}_k &= h_k(\hat{\mathbf{x}}_k^-, 0) + \left.\frac{\partial h_k}{\partial \mathbf{x}}\right|_{(\hat{\mathbf{x}}_k^-, 0)} (\mathbf{x}_k - \hat{\mathbf{x}}_k^-) + \left.\frac{\partial h_k}{\partial \mathbf{v}}\right|_{(\hat{\mathbf{x}}_k^-, 0)} \mathbf{v}_k \\ &= h_k(\hat{\mathbf{x}}_k^-, 0) + H_k(\mathbf{x}_k - \hat{\mathbf{x}}_k^-) + V_k\mathbf{v}_k \\ &= H_k\mathbf{x}_k + [h_k(\hat{\mathbf{x}}_k^-, 0) - H_k\hat{\mathbf{x}}_k^-] + V_k\mathbf{v}_k \\ &= H_k\mathbf{x}_k + \tilde{\mathbf{z}}_k + \tilde{\mathbf{v}}_k \end{aligned} \tag{23}$$

where $H_k$, $V_k$, $\tilde{\mathbf{z}}_k$ and $\tilde{\mathbf{v}}_k$ are defined by the above equation. The system (22) and measurement (23) now become linear and the PO-KF can be applied to obtain the PO-EKF for the robot localization as follows:

- The time update equations at prediction phase:

$$\begin{aligned} \hat{\mathbf{x}}_k^- &= f_{k-1}(\hat{\mathbf{x}}_{k-1}^+, \mathbf{u}_{k-n-1}, \mathbf{0}) \\ P_k^- &= A_k P_{k-1}^+ A_k^T + W_k Q_{k-1} W_k^T \end{aligned} \tag{24}$$

- The data update equations at correction phase:

$$
\begin{aligned}
M_* &= \prod_{i=1}^{m} A_{k-i}(I - K_{k-i}H_{k-i}) \\
K_k &= M_* P_s^- H_s^T (H_s P_s^- H_s^T + V_s R_s V_s^T)^{-1} \\
\hat{\mathbf{x}}_k^+ &= \hat{\mathbf{x}}_k^- + K_k[\mathbf{z}_k^* - h(\hat{\mathbf{x}}_s^-, \mathbf{0})] \\
P_k^+ &= P_k^- - K_k H_k P_s^- M_*^T
\end{aligned}
\tag{25}
$$

## IV. SIMULATIONS AND EXPERIMENTS

In order to evaluate the efficiency of the PO-EKF for the localization of networked mobile robot, simulations and experiments have been carried out.

### *1. Simulations*

Simulations are conducted in MATLAB with parameters extracted from a real networked robot system. The robot is a two wheeled, differential-drive mobile robot with the kinematics described in section II. The radius of two wheels is 0.05m and the distance between the wheels is 0.51m. The input noise is modeled as being proportional to the angular speed $\omega_{L,k}$ and $\omega_{R,k}$ of the left and right wheels, respectively. Thus, the variances equal to $\delta\omega_L^2$ and $\delta\omega_R^2$, where $\delta$ is a constant with the value 0.01 determined by experiments. The input-noise covariance matrix $Q_k$ is defined as:

$$
Q_k = \begin{bmatrix} \delta\omega_{R,k}^2 & 0 \\ 0 & \delta\omega_{L,k}^2 \end{bmatrix}
\tag{26}
$$

In the simulations, it is supposed that the robot has a sensor system that can directly measure the robot position and orientation in the motion plane. This measurement is suffered from a Gaussian noise with zero mean and the covariance:

$$
R_k = \begin{bmatrix} 0.01 & 0 & 0 \\ 0 & 0.01 & 0 \\ 0 & 0 & 0.018 \end{bmatrix}
\tag{27}
$$

Remaining parameters for the implementation of the PO-EKF are retrieved from the state-space model of the robot (3) as follows:

$$
A_k = \left.\frac{\partial f_k}{\partial \mathbf{x}}\right|_{(\hat{\mathbf{x}}_k^+, \mathbf{u}_k, \mathbf{0})} = \begin{bmatrix} 1 & 0 & -T_s v_c \sin\hat{\theta}_k^+ \\ 0 & 1 & T_s v_c \cos\hat{\theta}_k^+ \\ 0 & 0 & 1 \end{bmatrix}
\tag{28}
$$

$$
W_k = \left.\frac{\partial f_k}{\partial \mathbf{w}}\right|_{(\hat{\mathbf{x}}_k^+, \mathbf{u}_k, \mathbf{0})} = \begin{bmatrix} T_s \cos\hat{\theta}_k^+ & 0 \\ T_s \sin\hat{\theta}_k^+ & 0 \\ 0 & T_s \end{bmatrix}
\tag{29}
$$

$$
H_k = V_k = I
\tag{30}
$$

The Internet time delay *n(k)* in general is described as follows:

$$
\begin{aligned}
n(k) &= \sum_{i=0}^{N}\left[\frac{l_i}{C} + t_i^R + t_i^L(k) + \frac{M}{b_i}\right] \\
&= \sum_{i=0}^{N}\left(\frac{l_i}{C} + t_i^R + \frac{M}{b_i}\right) + \sum_{i=0}^{N} t_i^L(k) \qquad (31) \\
&= d_R + d_L(k)
\end{aligned}
$$

where $l_i$ is the $i^{th}$ length of link; $C$ is the speed of light; $t_i^R$ is the routing speed of the $i^{th}$ node; $t_i^L(k)$ is the delay caused by the $i^{th}$ node's load; $M$ is the amount of data; $b_i$ is the bandwidth of the $i^{th}$ link; $d_R$ is a term which is independent of time; and $d_L(k)$ is a time-dependent term. Because of the term $d_L(k)$ it is impossible to predict the internet time delay at every instant. In simulations and experiments, it is more efficient to measure the time delay at each sampling rate and use it for the localization. The measurement can be taken by adding timestamp to each sending message and performing the clock synchronization. Figure 3 shows the time delay measured in by our system in an experimental with the Internet.

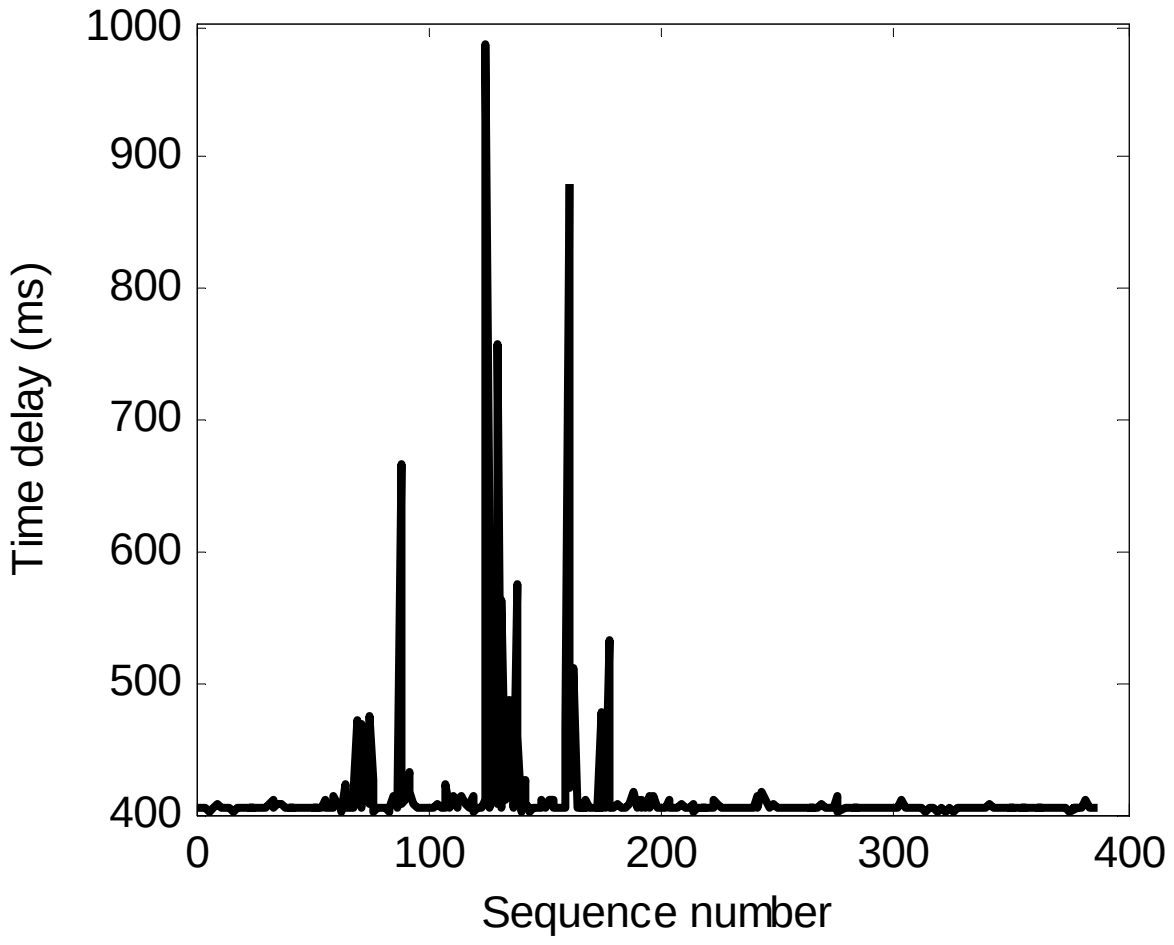

Figure 3: Time delay of the Internet measured in an experiment.

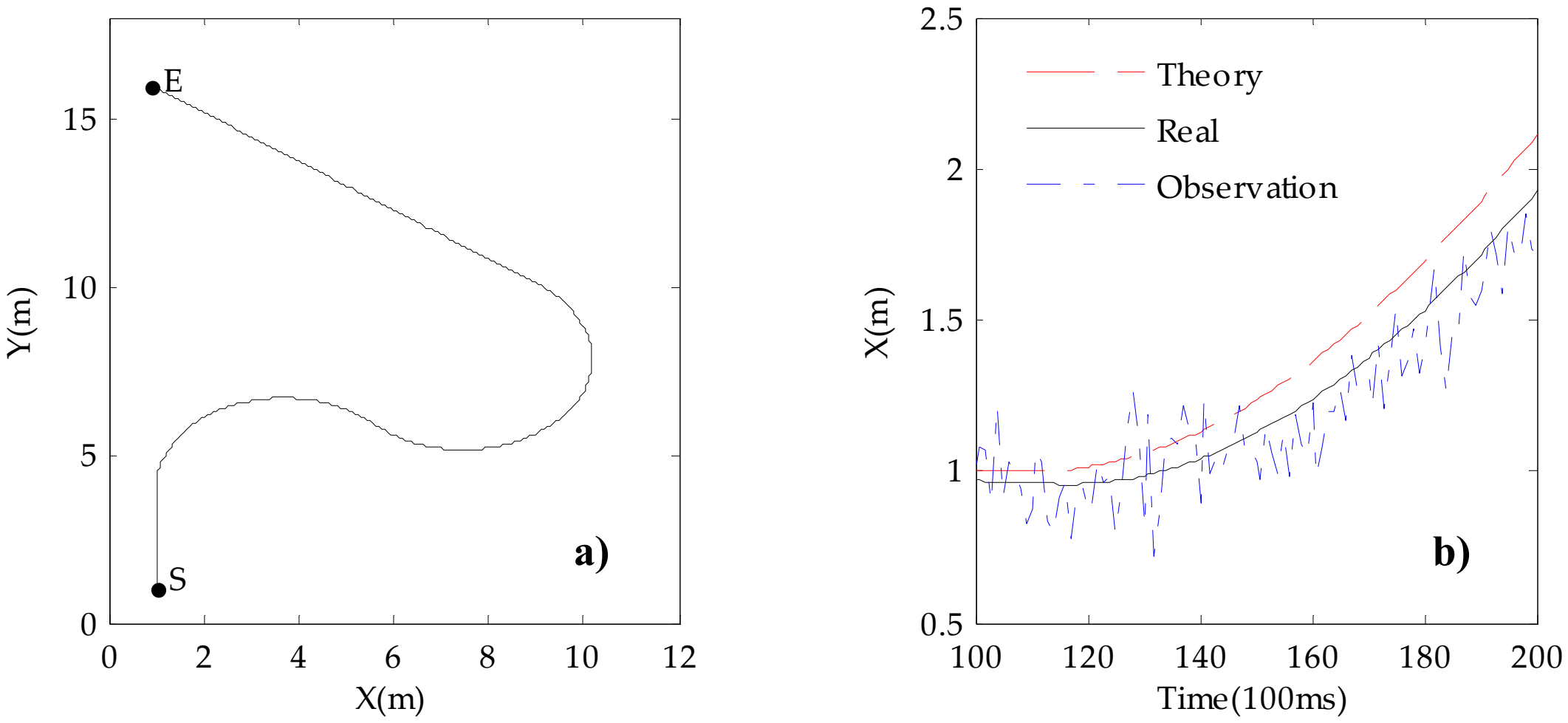

Figure 4: Trajectories of the robot in simulations

a) theory trajectory of the robot

b) A comparison between the theory, the real and the measurement trajectories

In evaluating the localization algorithm, the simulator constructs an arbitrary trajectory for the robot employing the dead-reckoning method, called *theory trajectory* (Figure 4a). Due to the input disturbance and communication delay, the robot actually follows a different path namely *real trajectory*. Measurements are performed to generate an *observation trajectory*. This trajectory suffers from the measurement noise and communication delay. Figure 4b shows trajectories in horizontal direction in a same coordinate. For the convenience of view and comparison, only 100 samples are displayed.

The localization is performed under two configurations: the normal extended Kalman filter (EKF) and the past observation-based extended Kalman filter (PO-EKF) to create *estimated trajectories*. Results for the scenario in which the control inputs and measurements are respectively delayed 300ms ($n$=3) and 400ms ($m$=4) are shown in Figure 5a. The deviation between the estimated trajectories and the real trajectory is derived in Figure 5b. It is recognizable that the PO-EKF introduces smaller deviation than the EKF. Having similar results, Figure 6 describes the vertical deviation between the estimated and the real trajectory in another simulation in which the robot follows a sinusoidal path.

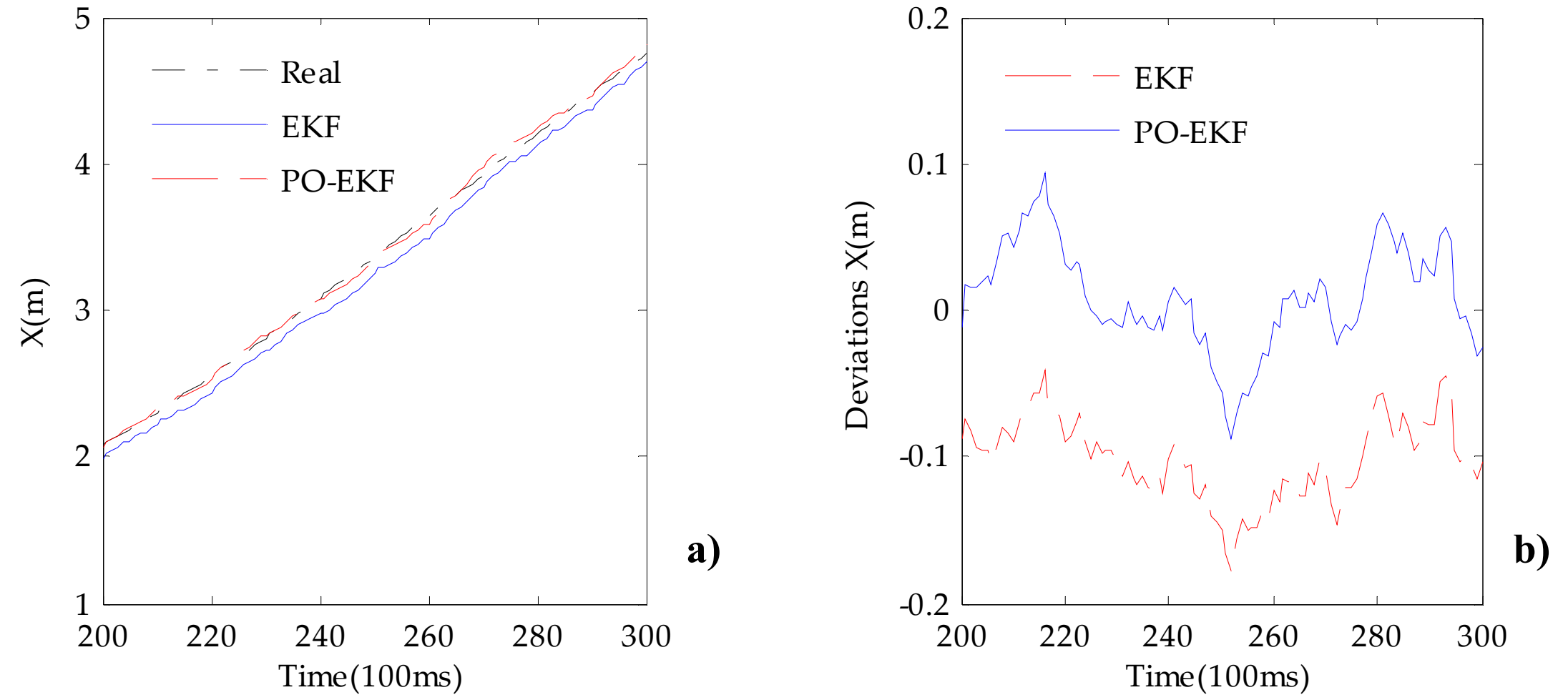

Figure 5: Comparison between the estimated trajectories using EKF and PO-EKF with $n$=3, $m$=4
a) Trajectories in the motion plan b) Deviations between the estimated and the real trajectories in the X direction

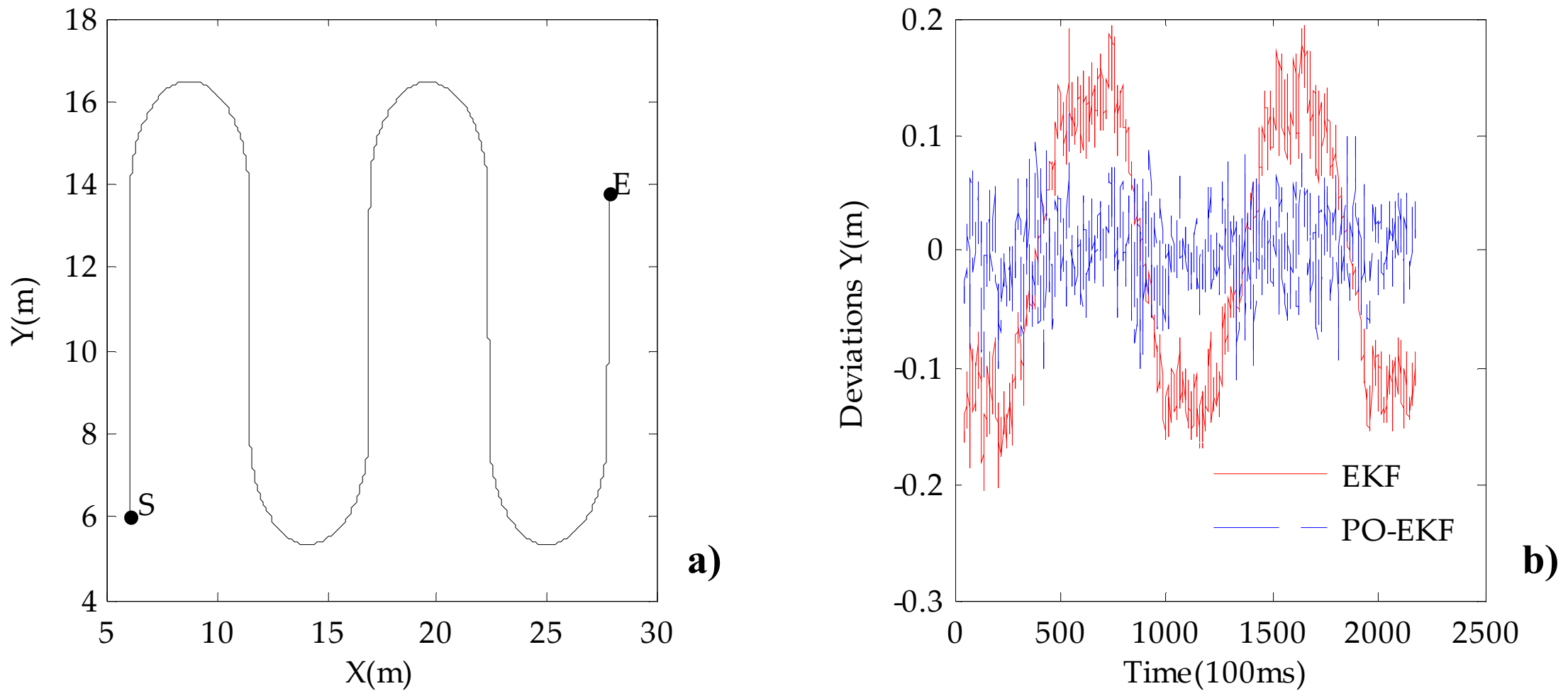

Figure 6: Comparison between estimated trajectories using the EKF and PO-EKF with $n$=2, $m$=4
a) Trajectories in the motion plan b) Deviations between the estimated and the real trajectories in the Y direction

## 2. Experiments

Experiments are carried out in a real networked mobile robot system (Figure 7). The robot is a Multi-Sensor Smart Robot (MSSR) developed at our laboratory. It contains basic components for motion control, sensing, navigation (Figure 8). The communication environment is the Internet and the 3G network is used. The remote controller is written in Visual C++ and implemented in a laptop computer. The time delay between the controller and the robot is determined by periodically transmitting probe messages. More details of the system can be referred from our previous work [16].

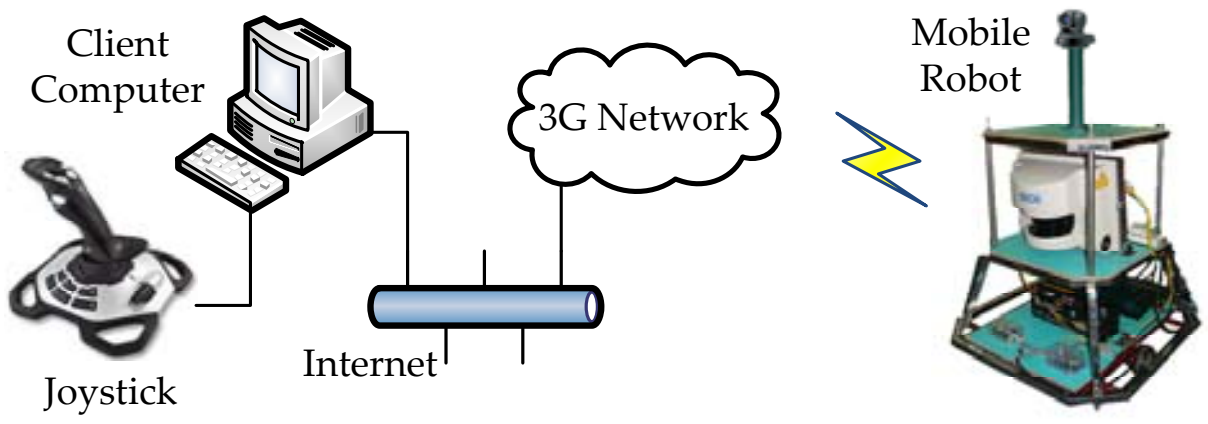

Figure 7: Hardware configuration of the Internet-based Robot system

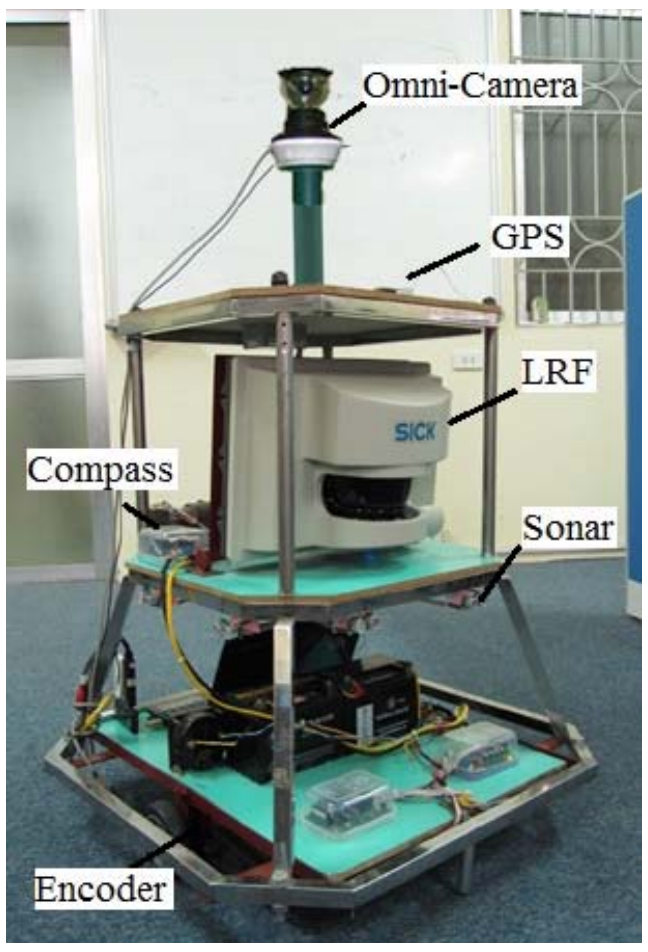

Figure 8: The networked mobile robot

In experiments, the robot is controlled in the manual mode to follow predefined paths. The localization is performed at two ends: the robot itself as no Internet and the remote controller with delayed data. Figure 9a shows the localization results in which the robot and the controller are connected to a local Internet service provider. The average time delay measured in this case is 483ms. The dot-dashed line presents the localization with no delay data while the dashed and the continuous lines respectively describe the results of the normal EKF and the PO-EKF algorithms with delayed data. Deviations between the trajectories with delayed data and the trajectory with no delay data are shown in Figure 9b,c,d. It is recognizable that the PO-EKF-based estimation is closer to the no-delay estimation than the EKF-based one. In another experiment, a VPN connection to a server located in the United States is utilized to simulate a far distance communication between the controller and the robot. Results show that average time delay is 773ms and the PO-EKF is able to compensate a certain amount of the time delay (Figure 10).

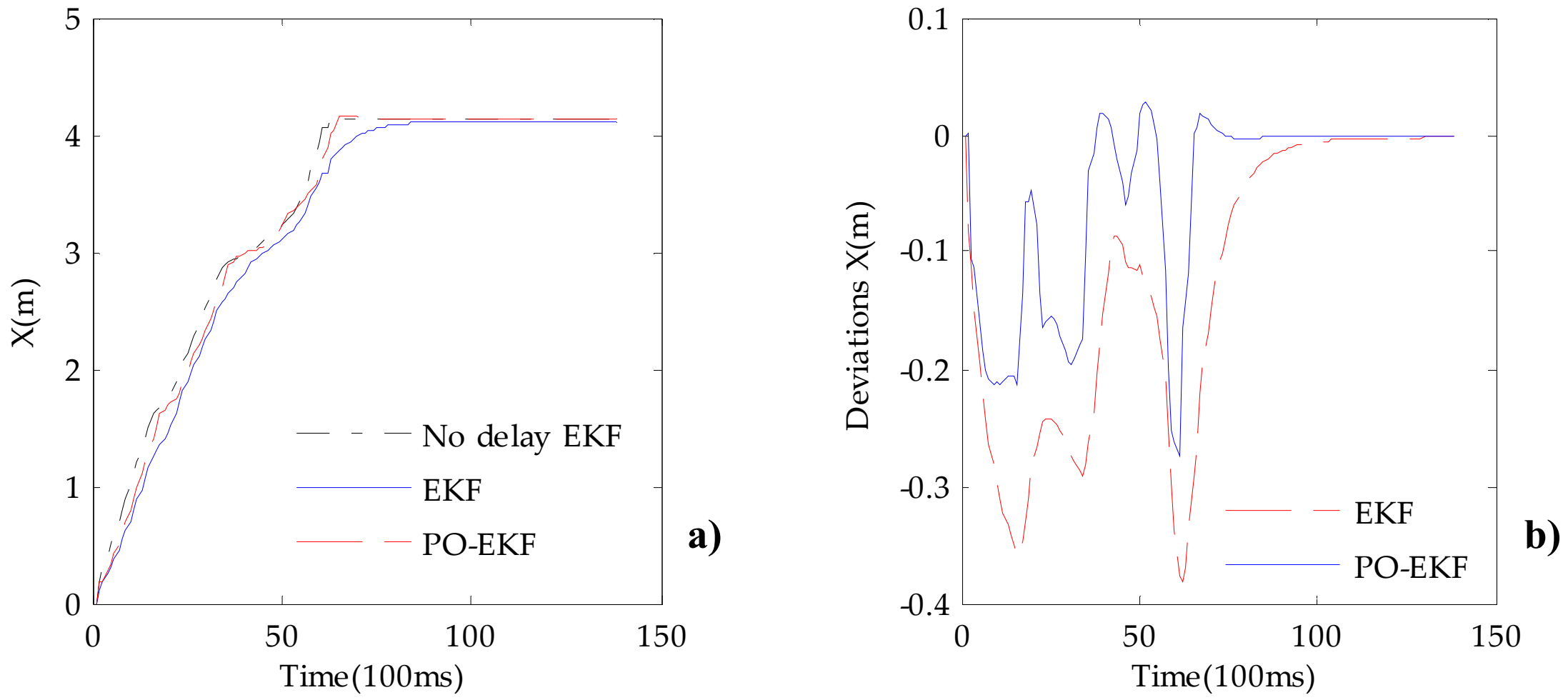

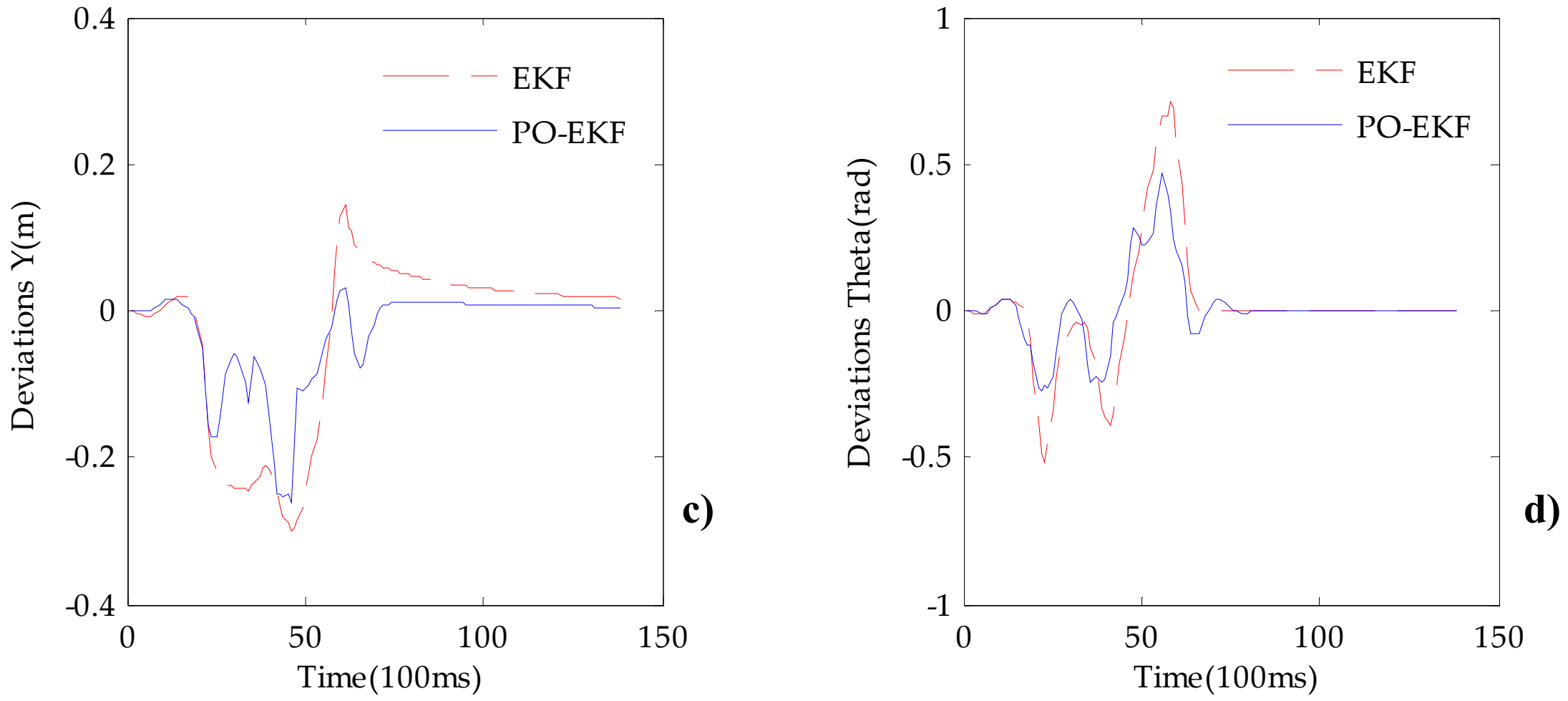

Figure 9: Estimated trajectories and their deviations with domestic Internet connections
a) Estimated trajectories
b) Deviations between the EKF and PO-EKF with the no-delay estimation in X direction
c) Deviations between the EKF and PO-EKF with the no-delay estimation in Y direction
d) Deviations between the EKF and PO-EKF with the no-delay estimation in orientation

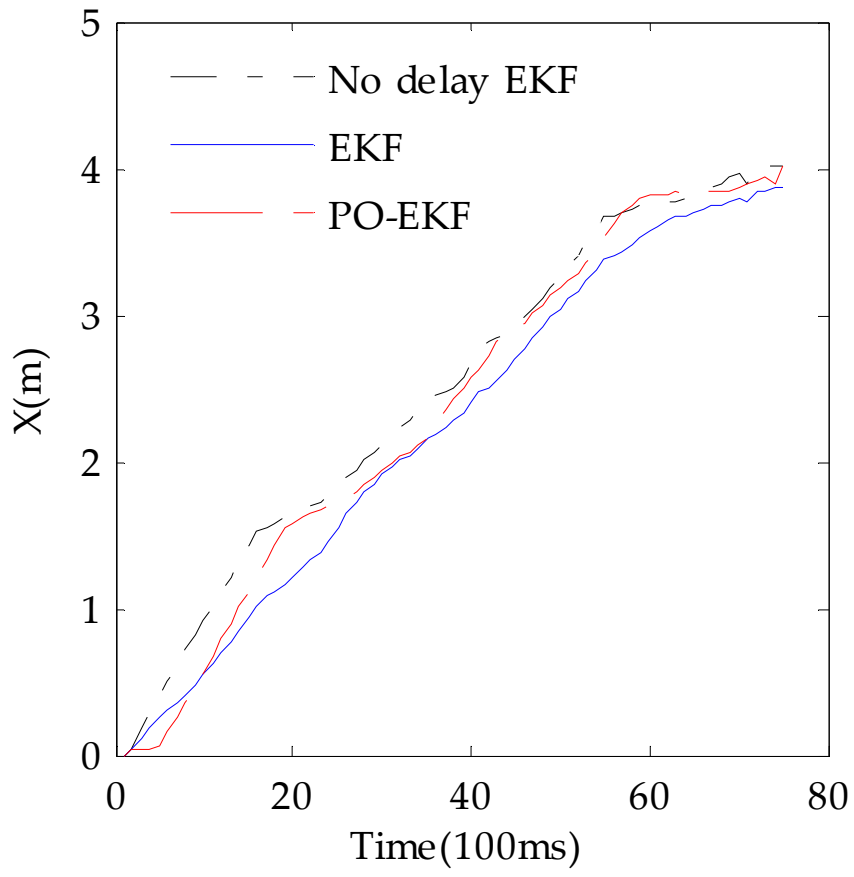

Figure 10: Estimated trajectories with a VPN connection to the United States

## V. CONCLUSION

In this paper, the localization problem of a mobile robot is investigated in the presence of Internet induced delay. A new state estimation algorithm namely PO-EKF is proposed in which the knowledge of system kinematics and the information of delayed measurements are combined. The optimality of the combination is theoretically analyzed and proven. A number of simulations and experiments have been conducted and the results confirm that the proposed algorithm is able to compensate the time delay caused by the Internet and the accuracy of the localization is sufficient for navigation tasks. In future work, other problems induced by the Internet such as the data loss and the out-of-order packet arrival will be addressed.

## ACKNOWLEDGMENT

This work was partly supported by the CN.12.15 project.

## REFERENCE

1. K. Goldberg and R. Siegwart, "Beyond Webcams: An Introduction to Online Robots", MIT Press, 2002.
2. E. Paulos and J. Canny, "Delivering real reality to the World Wide Web via telerobotics," Proceedings of the 1996 IEEE International Conference on Robotics and Automation, 1996.
3. D. Wang, J. Yi, D. Zhao and G. Yang, "Teleoperation System of the Internet-based Omnidirectional Mobile Robot with A Mounted Manipulator," Proceedings of the 2007 IEEE International Conference on Mechatronics and Automation, 2007.
4. Peter X. Liu, Max Q.-H. Meng, Polley R. Liu, and Simon X. Yang, "An End-to-End Transmission Architecture for the Remote Control of Robots Over IP Networks," IEEE/ASME transactions on mechatronics, Vol. 10, No. 5, 2005.
5. Alberto Sanfeliu, Norihiro Hagita, Alessandro Saffiotti, "Network robot systems", J. Robotics and Autonomous Systems 56 (2008) 793–797, Elsevier, 2008.
6. M. Shiomi, T. Kanda, H. Ishiguro, N. Hagita, "Interactive humanoid robots for a science museum", IEEE Intelligent Systems 22 (2) 25–32, 2007.
7. Ebastian Thrun, Dieter Foxb, Wolfram Burgardc, Frank Dellaerta, "Robust Monte Carlo localization for mobilerobots", J. Artificial Intelligence, Volume 128, Issues 1–2, p. 99–141, Elsevier, 2001.
8. Stephen Se, David Lowe, Jim Little, "Mobile Robot Localization and Mapping with Uncertainty using Scale-Invariant Visual Landmarks", The International Journal of Robotics Research, vol. 21, no. 8, p.735-758, 2002.
9. M. Betke, L. Gurvits, "Mobile robot localization using landmarks", IEEE Transactions on Robotics and Automation, Volume 13 , Issue 2, p.251 – 263, 1997.
10. S. Se, D. Lowe, J. Little, "Vision-based mobile robot localization and mapping using scale-invariant features", Proceedings 2001 ICRA. IEEE International Conference on Robotics and Automation, 2001.
11. Kuk-Hyun Han, Yong-Jae Kim, Jong-Hwan Kim and Steve Hsia, "Internet Control of Personal Robot between KAIST and UC Davis", Proceedings. ICRA '02. IEEE International Conference on Robotics and Automation, 2002.
12. Jong-Hwan Kim, Kuk-Hyun Han, Shin Kim and Yong-Jae Kim, "Internet-Based Personal Robot System using Map-Based Localization", Proceedings of the 32nd ISR (International Symposium on Robotics), 2001.
13. K. Han, S. Kim, Y. Kim, J. Kim, "Internet Control Architecture for Internet-Based Personal Robot", J. Autonomous Robots 10, 135–147, 2001.
14. Dan Simon, "Optimal State Estimation: Kalman Filter, H infinity, and Nonlinear Approaches", John Wiley & Sons Publication, 2006.
15. Greg Welch and Gary Bishop, "An Introduction to the Kalman Filter", Proceedings of SIGGRAPH, 2001.
16. P. M. Duong, T. T. Hoang, N. T. T. Van, D. A. Viet and T. Q. Vinh, "A Novel Platform for Internet-based Mobile Robot Systems", 7th IEEE Conference on Industrial Electronics & Applications (ICIEA), Singapore, 2012.